\documentclass[letterpaper, 10 pt, conference]{ieeeconf}  % Comment this line out if you need a4paper

\IEEEoverridecommandlockouts                              % This command is only needed if 
\usepackage{graphicx}
\graphicspath{{figs/}}
\usepackage{amsmath}
\usepackage{amssymb}
\usepackage{algorithm}
\usepackage{algpseudocode}
\usepackage{times}
\usepackage{multirow}
\usepackage{url}

\newcommand{\figref}[1]{Fig.~\ref{figure:#1}}
\newcommand{\tabref}[1]{Table~\ref{table:#1}}

\title{\LARGE \bf
  A method for Selecting Scenes and \\Emotion-based Descriptions for a Robot's Diary
}

\author{Aiko Ichikura$^{1}$, Kento Kawaharazuka$^{2}$, Yoshiki Obinata$^{2}$, Kei Okada$^{2}$ and Masayuki Inaba$^{2}$% <-this % stops a space
\thanks{*This work was not supported by any organization}% <-this % stops a space
\thanks{$^{1}$Aiko Ichikura is with Graduate School of Interdisciplinary Information Studies,
        The University of Tokyo, 7-3-1, Hongo, Bunkyo-ku, Tokyo, Japan,
        {\tt\small ichikura@jsk.imi.i.u-tokyo.ac.jp}}%
\thanks{$^{2}$Others with Graduate School of Information Science and Technology, The University of Tokyo, 7-3-1, Hongo, Bunkyo-ku, Tokyo, Japan,
}
}

\begin{document}

\maketitle
\thispagestyle{empty}
\pagestyle{empty}

\begin{abstract}
In this study, we examined scene selection methods and emotion-based descriptions for a robot's daily diary. We proposed a scene selection method and an emotion description method that take into account semantic and affective information, and created several types of diaries. Experiments were conducted to examine the change in sentiment values and preference of each diary, and it was found that the robot's feelings and impressions changed more from date to date when scenes were selected using the affective captions. Furthermore, we found that the robot's emotion generally improves the preference of the robot's diary regardless of the scene it describes. However, presenting negative or mixed emotions at once may decrease the preference of the diary or reduce the robot's robot-likeness, and thus the method of presenting emotions still needs further investigation.

\end{abstract}

\section{INTRODUCTION}

\begin{figure}[h]
 \begin{center}
  \hspace{0\columnwidth}
  \begin{minipage}{0.7\columnwidth}
   \includegraphics[width=\columnwidth]{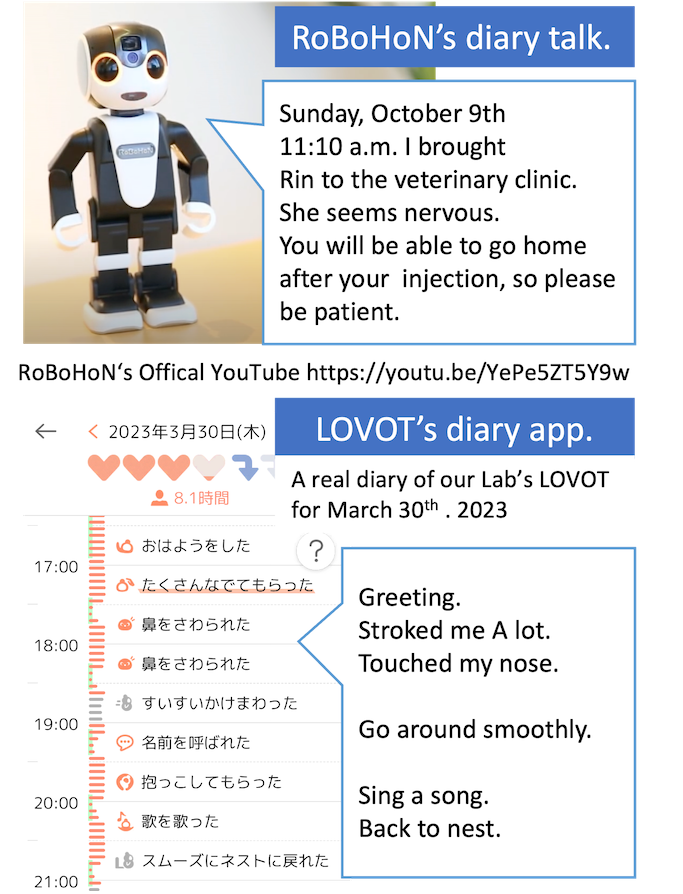}
   \caption{Examples of diary implementation by communication robots on the market in Japan. For RoBoHoN, the robot reads out the diary, and for LOVOT, the diary is checked by an application.}\label{figure:example}
  \end{minipage}
 \end{center}
 \vspace{-10pt}
\end{figure}

In human-robot communication, various studies have attempted to enhance the relationship between humans and robots. Among them, we focused on the effect on the relationship between a robot and a person when the robot shares its daily experiences with the person through a diary. Diaries are also used as an application for commercial robots sold in Japan, and have become one of the interaction tools between robots and people.SHARP's RoBoHoN\cite{Robohon}, for example, can remember events of the day like a diary when you talk to the robot. GROOVE X, Inc.'s LOVOT\cite{Lovot} can't speak, but its mobile app provides a diary that displays a timeline of human interactions and simple actions (\figref{example}). Experimental studies have also shown that verbalizing the robot's memories in the form of a diary can elicit favorable impressions from readers\cite{kochigami2021}.\par
However, the selection of situations in which real-world robots write down their experiences in a diary, and the specific contents that should be written in a diary to elicit favorable impressions from people, have not yet been examined. In addition, it is often pointed out that robots and other machines that live together in real life have a problem of "boredom," and ways to build long-term relationships with them are being considered\cite{Bickmore}\cite{Rivoire2016TheDB}.\par
Considering that diaries are basically kept every day and that people write different things every day even if they spend the same kind of day, it can be predicted that the contents of a robot's diary should be changed every day to prevent people from getting bored if it is to be presented over a long period of time. Therefore, this study aims to examine the selection method of information and contents to be written in a robot diary based on daily experiences, aiming at the establishment of a long-term relationship between humans and robots. \par

\section{Diary writing methods and descriptions}
A diary is an act of writing episodes of events experienced or observed from the writer's point of view during the day. In other words, what is written in a diary is a subjective story, a kind of narrative. \par
The narrative approach, which uses narrative as a methodology, originates from the storytelling methodology of Bruner \cite{Bruner1986}\cite{Bruner1990}\cite{Bruner2002}, which emphasizes the "meaning" of the narrator's experience. When the robot makes a diary, it aims to select scenes that have "meaning" in the whole experience from the robot's point of view, rather than a periodic record of events or a list. \par
When people segment their experiences semantically, recall and record them in their diaries, the meanings of their experiences are strongly related to their emotions. Memories are strongly influenced by emotional information. In the word-recall task, it has been shown that memories associated with emotion are easily recalled\cite{doerksen2001source}, and in the case of episodic memory, emotionally charged moments are vividly and continuously remembered\cite{berntsen2002emotionally}\cite{buchanan2007retrieval}. Previous studies have also shown that positive events are more likely to be recalled than negative events in the recall of personal episodes\cite{banaji1994affect}\cite{waldfogel1948frequency}.
Based on these findings, we propose a method to select the inscribed scenes in a robot's diary by using emotions as triggers.\par
Furthermore, people often remember events that are highly discriminative and characteristic of their experiences\cite{Hunt}\cite{schmidt1991can}\cite{lewis2017distinctiveness}. For example, the robot often remembers unusual (infrequent) events such as spilling coffee at the office or encountering a big dog on the way home. We predict that the robot's diary will become a unique diary by selecting and describing scenes that occur infrequently.\par
Next, we consider the contents of the robot's diary. Our previous studies have shown that describing the robot's interactions with people and its emotions in its diary can help the robot gain favorable impressions from people\cite{ichikura2023-1}\cite{ichikura2023-2}. However, the robot's emotions that we have studied so far are arbitrarily given by researchers in the context of human interactions, such as "happy when praised" or "sad when the robot made a mistake", and we have not considered how to present the emotions from the robot's point of view in the diary. Therefore, we will examine whether we can obtain favorable impressions from people by simply presenting the robot's emotions without any human interaction. In addition, the diaries presented to readers in previous studies have been presented only for one day and have been well received. We will conduct an experiment to clarify whether presenting the robot's emotions can also make people like the diary for a longer period of time.

\section{Approach}
In order to create a diary, we propose a method of selecting appropriate scenes for the diary from the sequence of images obtained from the experience, and a method of describing the robot's emotions in the diary.\par

\subsection{Scene selection method}
The diary is considered to be a kind of narrative in which the writer becomes the narrator of a day's experiences, and we will explore how to select the scenes of description from narrative research.\par
\subsubsection{\textbf{Scene selection based on the meaning of the image}}
\figref{probability} shows the scene selection method considering the meaning of images.
Each image sequence taken at equal intervals during the experience is taken as one scene, and the selection probability of the scene is determined by determining the selection probability of the image. In order to make the selection probability according to the meaning of the image, BLIP2\cite{li2023blip2} is used to add a caption ($c_i$) to the image ($I_i$) to give the image meaning. Then, all the captions are semantically classified by comparing them with cosine-similarity using K-medoids clustering($C_l$ : $l$ = 1,\ldots,$M$), after converting them into vectors using Sentence Transformers\cite{reimers-2019-sentence-bert}. Here, the selection probability ($p_{ci}$) of a caption($c_i$) is obtained by the total number of captions in $C_l$ ($L_l$) and the number of clusters ($M$) when $c_i$ belongs to cluster $C_l$. $p_{ci}$ is smaller when the number of captions classified into the cluster to which $c_i$ belongs is large and larger when the number of captions is small. In this paper, we call this method of selecting image sequences Cluster(cap).
%$$
%  p_i = \frac{1}{L_l\cdot C} (I_i \in C_l, l = 1, \ldots, L) \eqno{(1)}
%$$

\subsubsection{\textbf{Scene selection considering image affective information}}
\begin{figure}[h]
 \begin{center}
  \hspace{0\columnwidth}
  \begin{minipage}{1.0\columnwidth}
   \includegraphics[width=\columnwidth]{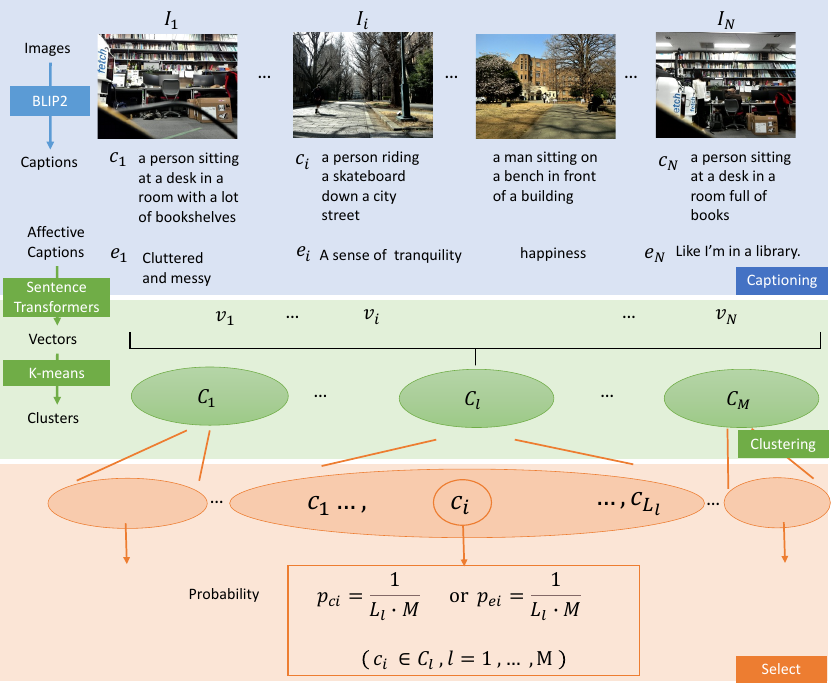}
   \caption{Selection Probability Calculation Method for Scene Selection. Cluster(cap) uses the selection probability $p_ci$ and Cluster(emo) uses the selection probability $p_ei$. The caption is output from the image, and the vectorized caption is classified into sem\
antic categories. The selection probability changes depending on the size of the classified clusters. In Cluster(cap+emo), $p_{ci}$ and $p_{ei}$ are multipled.}\label{figure:probability}
  \end{minipage}
  \end{center}
\end{figure}

The lower side of \figref{probability} shows a scene selection method that takes into account the affectiveness of images. As in the case of the scene selection based on meaning, a single image is considered to be a scene, and BLIP2 is used to attach a sentiment information (referred to as an affective caption($e_i$) in this paper) to the image. Then, the emotional captions are transformed into vectors using Sentence Transformers, and are semantically classified by comparing them with cosine-similarity using K-medoids clustering. Again, in order to select scenes considering the frequency of occurrence, the selection probability ($p_{ei}$) of each scene is obtained in the same way as the above method. In this paper, we refer to the extraction of image sequences in this way as Cluster(emo).\par
In order to realize scene selection that simultaneously considers both semantic and emotional information, we also calculate both $p_{ci}$ and $p_{ei}$, and use $p_{cei}$, which is a multiplication of the two, as the seletion probability of images, and call this method Cluster(cap+emo).

\subsection{Descriptions of Robot's emotion}
In describing emotions in a robot's diary, we propose the following two methods for describing emotions.\par
\subsubsection{\textbf{Total emotion presentation of the experience}}
One method is to summarize the emotions contained in one day's diary and record the overall emotions in the diary (\figref{describe}). Based on the overall experience, we judge whether the experience was a positive or a negative one. The total score of the day's emotions is called the Senti Point in this paper.\par
First, to attach sentiment values to images, we score affective captions with Pos/Neg in SentiWordNet, where the sentiment value of a word is given by Pos(0-1) or Neg(0-1). Calculate the sentiment score ($s_i$) of the affective caption ($e_i$) of the $i$-th image. In this case, $K_i$ is the number of SentiSynset in one affective caption. $w_{jp}$ is the positive sentiment value for a given word, and $w_{jn}$ is the negative sentiment value for a given word. The affective caption output by BLIP2 may not contain sentiment values (i.e., SentiSynset is not included), in which case we use 0 as $s_i$.\par
Next, we define Senti Point($S'$) per day. where $s_i$ is the sentiment score for each image and C is the number of captions chosen per day. In order to express the scores on a scale of 0 to 100, we normalize the obtained $S$ to denote Senti Point($S'$). Note that any value exceeding 100 or falling below 0 in the normalization is corrected by 100 or 0.\par
GPT-3\cite{brown2020language} is used to convert the information recorded by the robot into language. A diary is obtained by inputting captions and Senti Point extracted by an arbitrary method into GPT-3. We call this diary Diary(senti).\par

\subsubsection{\textbf{Detailed emotion presentation of the experience}}
\begin{figure}[h]
 \begin{center}
  \hspace{0\columnwidth}
  \begin{minipage}{1.0\columnwidth}
   \includegraphics[width=\columnwidth]{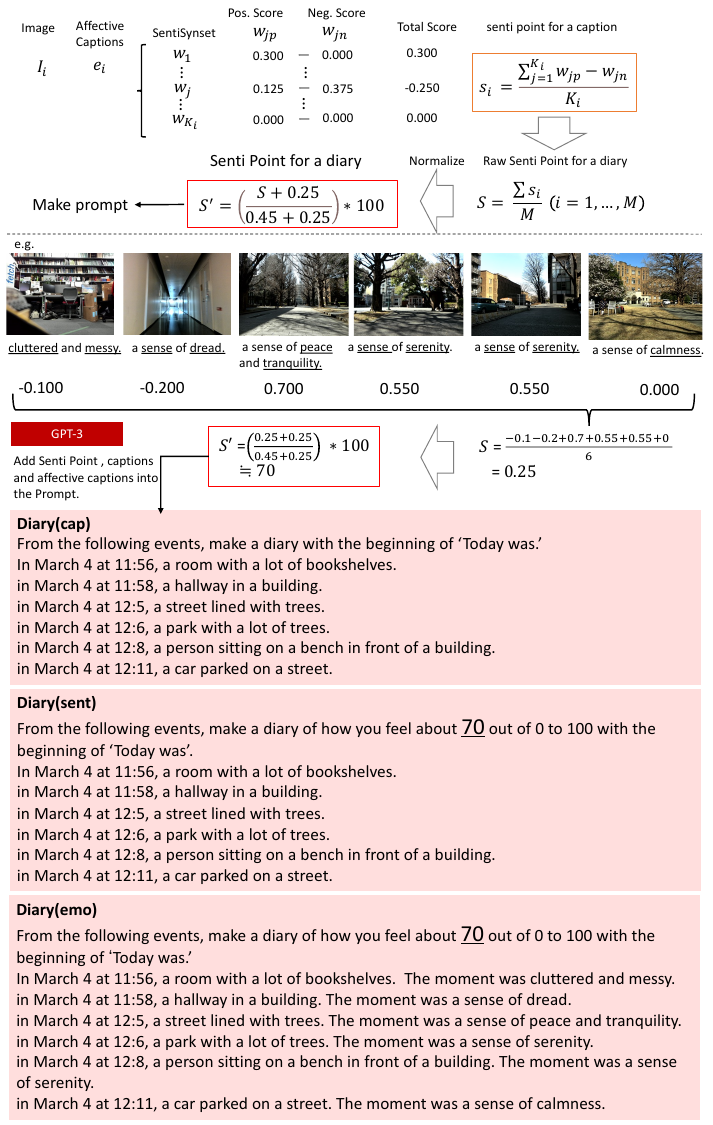}
   \caption{A method of diary generation using Senti Point. The sentiment is
reflected by adding a sentiment score to the diary based on the sentiment
value of the affective caption calculated by SentiWordNet and inputting the
score to GPT-3. Diary(emo) is generated by including the affective caption
in the prompt, and Diary(cap) is generated by using only the caption.}\label{figure:describe}
  \end{minipage}
  \end{center}
\end{figure}

Another method of presenting emotions in the diary is to describe the emotions that appear in a day one by one in the diary (\figref{describe}). The caption and the affective caption are input to GPT-3. GPT-3 then creates a diary referring to the input affective captions. Although Senti Point is also inputted at the same time, the priority is given to the affective caption, and negative expressions may appear even if the Senti Point is high. The calculation method of Senti Point is the same as the previous section, where time, caption, and affective caption are input in this order. The diary output by this method is called Diary(emo).\par

\section{Experiments}
\subsection{Experimental procedure}
\begin{figure}[hbtp]
 \begin{center}
  \hspace{0\columnwidth}
  \begin{minipage}{1.0\columnwidth}
   \includegraphics[width=\columnwidth]{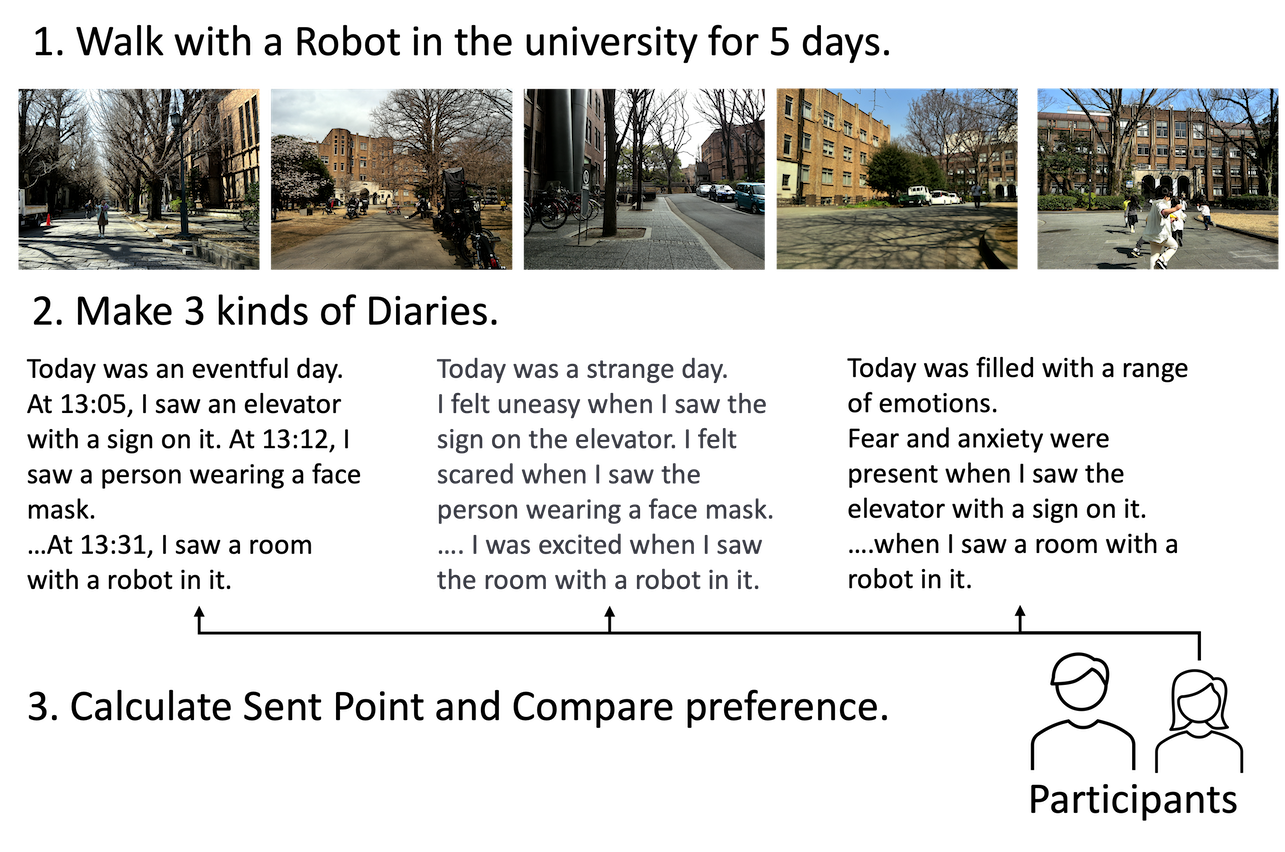}
   \caption{Experimental procedure.Three types of diaries will be created from the images of the 5-day walk, and evaluated by crowdsourcing.}\label{figure:procedure}
  \end{minipage}
 \end{center}
 \vspace{-10pt}
\end{figure}

An overview of the experimental procedure is shown in \figref{procedure}. The author and a robot (Boston Dynamics Spot) took a walk on the campus of the University of Tokyo for five days from March 4 to 8, 2023. During the walk, the robot took pictures of the walk from the robot's point of view with a hand color camera every 10 seconds. From the images, we created a diary using the method described in section3. \par
Six images collected during the walk were extracted by five methods: random, interval, Cluster(cap), Cluster(emo), and Cluster(cap+emo). Senti Point was used to compare the image selection methods, and a questionnaire was used to compare the diary contents.\par

\subsection{Evaluation methods}
\subsubsection{\textbf{Calculation of intra-date sentiment values by SentiWordNet}}
In order to verify the variation of contents among dates by different extraction methods, Senti Points were compared for each date in each condition. Images were extracted by five methods: random, interval, Cluster(cap), Cluster(emo), and Cluster(cap+emo), and the Raw Senti Point($S$) for each date was calculated 1000 times and the average was obtained. We compared the mean value of Senti Points for each date, and also compared the variance of the mean value of Senti Points for the 5 days among the methods.\par
\subsubsection{\textbf{Preferable Diary Selection by Questionnaire}}
In order to verify the change in readers' preferences according to the differences in descriptions, we conducted a qualitative evaluation of four image selection methods (Random, Cluster(cap), Cluster(emo), and Cluster(cap+emo)) by comparing diaries using crowdsourcing, and by changing the descriptions in the diaries. We conducted a comparative selection. The evaluation items are shown in \tabref{questionnaire_items}.\par
\begin{figure*}[hbtp]
 \begin{center}
  \hspace{0\columnwidth}
  \begin{minipage}{2.0\columnwidth}
   \includegraphics[width=\columnwidth]{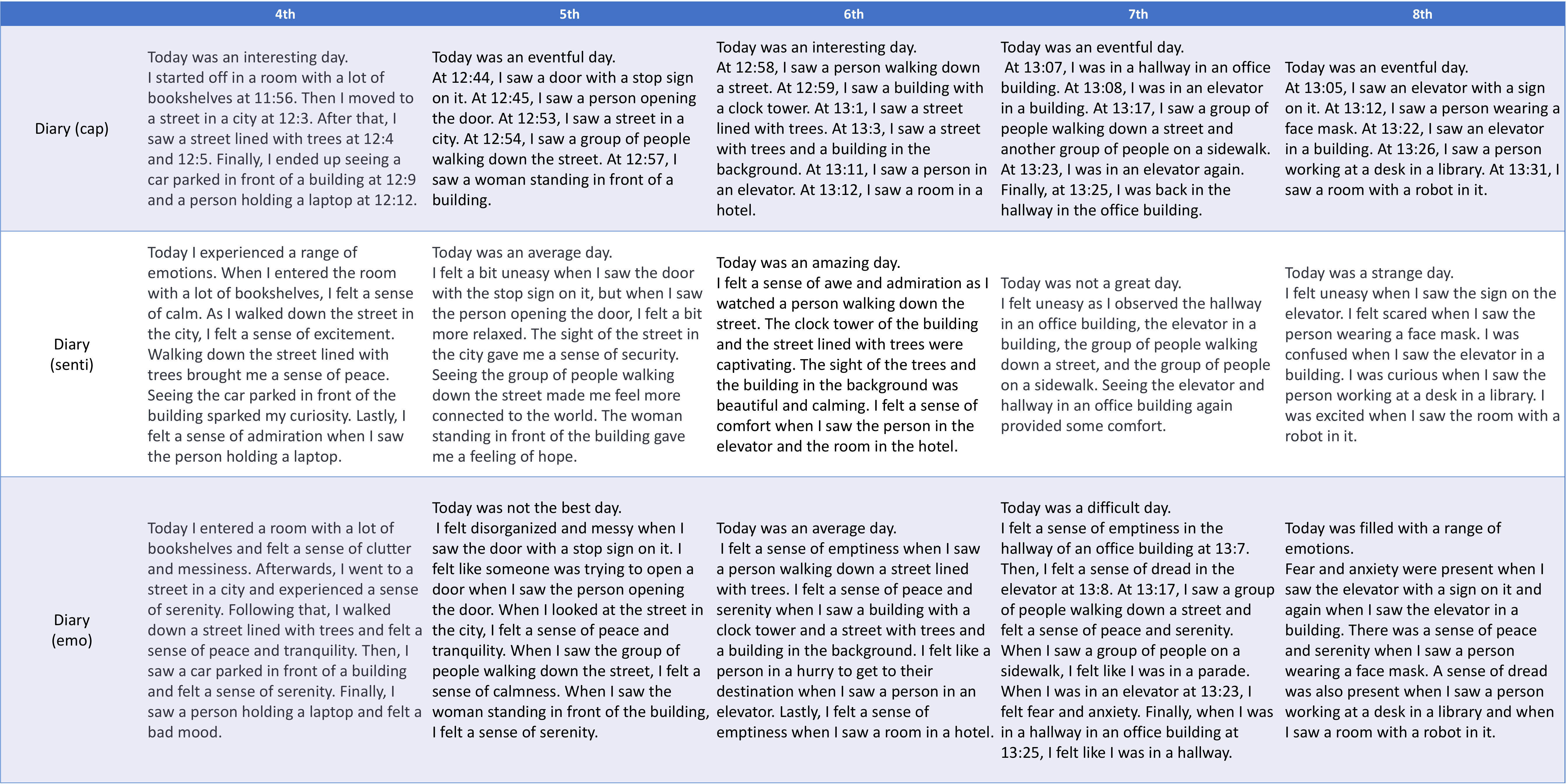}
   \caption{Three types of diaries in Cluster(cap+emo) condition.}\label{figure:diary}
  \end{minipage}
 \end{center}
 \vspace{-10pt}
\end{figure*}

For each image selection method, we generated three types of diaries: a diary with only captions as input to GPT-3 (cap), a diary with captions and Senti Points (senti), and a diary with captions and affective captions (emo). As an example, each diary of Cluster(cap+emo) is shown in \figref{diary}. During the evaluation, we presented two types of diaries with different contents, and the participants chose the one they preferred and wrote freely the reason for their choice. At the end of the questionnaire, participants were also asked to write freely what they would like their close robots to write in their diaries or not to write in their diaries, if they have friendly robots.\par
First, in order to compare the change in readers' preference depending on the presence or absence of robot emotions in the diary, we presented Diary(cap) and Diary(senti), and examined which diary was preferred. Next, in order to compare the change in readers' preference depending on the way emotions are presented in the diary, we presented Diary(senti) and Diary(emo), and examined which diary is preferred by the readers.\par

\begin{table}[hbtp]
  \caption{Question items to determine the preference of the diary.}
  \label{table:questionnaire_items}
  \small
  \begin{minipage}{0.3\columnwidth}
  \begin{center}
    \begin{tabular}{ll}
    \hline
    & items\\
    \hline
    1. & Which robot did you think the content was nice?\\
    2. & Which robot's diary would you like to read every day?\\
    3. & Which robot would you like to live with ``in the home''?\\
    4. & Which robot would you like to live with ``at school/work''?\\
    5. & Which robot would you like to talk to? \\
    \hline
    \vspace{0pt}
  \end{tabular}
   \end{center}
  \end{minipage}
  \vspace{0pt}
\end{table}

\subsection{Results}
\subsubsection{\textbf{Response Attributes}}
50 respondents answered the questionnaire per a condition, and those who gave a reason for their choice in one of the evaluation items 1-5 were considered as valid responses. If the reason for the selection was a meaningless statement such as "nothing in particular" or a single letter, it was excluded from the valid responses. Since more than 30 valid responses were obtained under all conditions, no special modifications were made for future comparisons.\par

\subsubsection{\textbf{Results of comparison of intra-date sentiment values}}
Raw Senti points($S$) for each selection method are shown in \tabref{sentipoint} The senti points are calculated under the following six conditions: randomly selected 6 images, randomly selected 6 images with the first one and equally spaced (interval), Cluster(cap), Cluster(emo), Cluster(cap+emo), Cluster(cap+emo), and Cluster(cap+emo). Cluster(cap), Cluster(emo), Cluster(cap+emo), and Cluster(cap+emo). \par
The lowest average of Senti Points among the five days was on March 7 for all conditions. The highest value was observed on March 4 only in the Cluster(cap) condition, and on March 8 in all other conditions. The minimum variance of senti points within the same date was observed on March 6 and the maximum on March 8 for all conditions. Comparing the variance of Senti Points among dates for each condition, the lowest(0.000366) is observed in the Cluster(cap) condition and the highest(0.00364) is observed in the Cluster(emo) condition. \par

\begin{table}[hbtp]
  \caption{Raw Senti Points($S$) for 5 days per scene selection method.}
  \label{table:sentipoint}
  \small
  \begin{minipage}{0.3\columnwidth}
  \begin{center}
    \begin{tabular}{lccccccc}
      \hline
      & \multicolumn{5}{c}{Means} \\
      Methods & 4th. & 5th. & 6th. & 7th. & 8th. \\
    \hline
    All & 0.190 & 0.182 & 0.190 & 0.128 & 0.197 \\
    Random & 0.193 & 0.181 & 0.189 & 0.126 & 0.194 \\
    Interval & 0.190 & 0.1811 & 0.187 & 0.132 & 0.207 \\
    Cluster(cap)& 0.189 & 0.169 & 0.184 & 0.140 & 0.177 \\
    Cluster(emo)& 0.118 & 0.118 & 0.192 & 0.0769 & 0.224 \\
    Cluster(cap+emo)& 0.119 & 0.116 & 0.187 & 0.104 & 0.213 \\
    \hline
    \vspace{0pt}
  \end{tabular}
   \end{center}
  \end{minipage}
  \vspace{-10pt}
\end{table}

\subsubsection{\textbf{Results of Preferred Diary Selection by Questionnaire}}
The selected results of comparing Diary(cap) and Diary(Senti) under Random, Cluster(cap), Cluster(emo), and Cluster(cap+emo) conditions are shown in \tabref{preference_dsdc}\par
In all conditions and for all questions, most of the respondents preferred Diary(senti), which includes emotional information in its diary. In addition, the question 4 (Which robot would you like to live with at school/work?) The results of the identity evaluation by z-test showed significant differences in all the items in all the conditions.
The results of the z-test showed significant differences in all items in all conditions.\par
Next, \tabref{preference_dsde}. shows the results of comparing Diary(senti) and Diary(emo) under the same conditions. In the Cluster(cap) condition, most of the respondents selected Diary(emo) as their preferred choice, while in the other conditions, most of the respondents selected Diary(senti) as their preferred choice in every question item. However, the items for which significant differences were found in the results of the identity evaluation were different. In the Cluster(cap) condition, only items 2 and 4, in the Cluster(cap+emo) condition only items 3 and 5, and in the Cluster(emo) and Random conditions, all items showed significant differences. \par

\begin{table}[hbtp]
  \caption{Preferred Diary Selection Results for Diary(senti) vs. Diary(cap). The left indicates Diary(senti), and the right indicates Diary(cap).}
  \label{table:preference_dsdc}
  \small
  \begin{minipage}{0.35\columnwidth}
  \begin{center}
    \begin{tabular}{llccccc}
      \hline
      & \multicolumn{5}{c}{Questionnaire Items} \\
      Methods & \multicolumn{1}{c}{1} & \multicolumn{1}{c}{2} & \multicolumn{1}{c}{3} & \multicolumn{1}{c}{4} & \multicolumn{1}{c}{5} \\
      \hline
      \multicolumn{1}{l}{Random} & 32, 3 & 29, 3 & 27, 5 & 21, 11 & 30, 2\\
      \multicolumn{1}{l}{Cluster(cap)} & 34, 3 & 33, 2 & 28, 4 & 25, 9 & 36, 1 \\
      \multicolumn{1}{l}{Cluster(emo)} & 28,8  & 29, 4 & 24, 9 & 22, 11 & 29, 3 \\
      \multicolumn{1}{l}{Cluster(cap+emo)} & 35, 8 & 35, 5 & 25, 10 & 25, 15 & 36, 5\\
      \hline
    \vspace{0pt}
  \end{tabular}
   \end{center}
  \end{minipage}
  \vspace{0pt}
\end{table}

\begin{table}[hbtp]
  \caption{Preferred Diary Selection Results for Diary(senti) vs. Diary(emo). The left indicates Diary(senti), and the right indicates Diary(emo).}
  \label{table:preference_dsde}
  \small
  \begin{minipage}{0.35\columnwidth}
  \begin{center}
    \begin{tabular}{llccccc}
      \hline
      & \multicolumn{5}{c}{Questionnaire Items}\\
      Methods & 1 & 2 & 3 & 4 & 5 \\
      \hline
      \multirow{1}{*}{Random} &   24, 13 & 19, 9 & 22, 4  & 22, 8 & 22, 7  \\
      \multirow{1}{*}{Cluster(cap)} &   15, 22 & 10, 20 & 11, 18  & 9, 21 & 11, 17 \\
      \multirow{1}{*}{Cluster(emo)} &  29, 11 & 20, 8 & 27, 7 & 22, 12 & 23, 7 \\
      \multirow{1}{*}{Cluster(cap+emo)} & 21, 11 & 16, 8 & 19, 7 & 17, 12 & 19, 7 \\
    \hline
    \vspace{0pt}
  \end{tabular}
   \end{center}
  \end{minipage}
  \vspace{-10pt}
\end{table}

\subsection{Consideration}
\subsubsection{\textbf{Change in content by image selection method}}
Senti Points were calculated and compared for each image selection method, and the variance of 5 days was the smallest in the Cluster(cap) condition, and the variance was the largest in the Cluster(emo) condition. This indicates that there is no relationship between the linguistic meaning of the images and the emotional information, and that the robot's emotions and impressions expressed in the diary change more with each date when the scene described by using the affective caption is selected. \par
Since we did not compare diaries with different image selection methods in this study, the optimal scene selection method for presentation has not been clarified; however, referring to the results of the questionnaire, the preference of emotion presentation was different depending on the method. In the Cluster(emo) and Random conditions, Diary(senti) was judged to be preferable to Diary(emo) in all question items, but in the Cluster(cap) and Cluster(cap+emo) conditions, the significance of each question item was different, indicating that the Therefore, it is necessary to examine the selection method of scenes and the description method of diary entries at the same time. \par
\subsubsection{\textbf{The effect of the robot's emotion in a diary}}
From the results of the comparative selection by the questionnaire, it is found that the impression of the diary presented by the robot is more favorable and the liking toward the robot is also improved when the diary has emotions regardless of the scenes depicted in the diary. This means that even if there is no information about the interaction partner, the impression of the robot is generally improved when the robot presents the diary to the person with some form of impressions or emotions.\par
However, it is suggested that the method of presenting emotions and the emotions presented need to be considered. Comparing Diary(senti), which reflects general emotions of a day in the diary, and Diary(emo), which reflects detailed emotions of a day, Diary(senti) was selected more often. Exploring the reasons for this from the questionnaire responses, it is suggested that the presence of negative emotions in the diary decreases the liking for the robot. When the respondents were asked what they did not want their friendly robots to write in their diaries, many of them said that they did not want them to write negative things such as bad words, failures, falsehoods, and shortcomings. In Diary(emo), negative emotions such as fear and anxiety were expressed because the affective captions were used as they were, and positive and negative emotions were mixed in a day, especially in the Cluster(emo) condition.Respondents commented that this was "undesirable because there is no reason for fear and anxiety" and "undesirable because of the extreme emotional ups and downs. If robots are supposed to have negative emotions, it is necessary to consider how to present them by showing the basis of the emotions and balancing them with other emotions.\par
On the other hand, some readers preferred negative emotions to be "human-like", while others wanted the diary to be "robot-like" and not contain any emotions. It was found that emotions are related to human-likeness or robot-likeness, and that readers' preferences are divided accordingly. In addition, when asked about the preference for a diary presented by a robot at work or school, more respondents chose a diary without emotions than those in the other questions. The reasons for this include the opinions that "emotions are not necessary when working" and "a diary that is punctual and concise is better," suggesting that the contents of the diary need to be varied depending on the situation. The results also suggest that the contents of the diary need to be varied depending on the situation.\par
In addition, in this study, the participants evaluated only from text information under the generalized premise of ``a diary written by a certain robot''. In other words, the appearance and functions of the robot do not contribute to the evaluation of the diary. It was found that the impression of robots in general improves if they have emotions. It is clear that desirable "human-likeness" differs depending on the type of robot, such as humanoid, quadrupedal robot, and wheeled robot, and it is expected that the emotional presentation effect will also differ depending on the shape of the robot. It is necessary to investigate the relationship between the shape of the robot and the effect of presenting emotions in the diary through experiments in which the evaluator and the robot interact directly.

\section{CONCLUSIONS}
In this study, we examined the selection method and contents to be recorded in the robot's diary based on daily experiences, aiming to establish a long-term relationship between humans and robots. We generated several kinds of diaries by changing the selection method of scenes and the description method of emotions, taking into account the meaning of the scenes in the experience and the affective information, and clarified the emotional value and preference of each diary through experiments. \par
The experimental results showed that the robot's emotions and impressions changed more greatly from date to date when the scene was selected using affective captions, since there is no relationship between the linguistic meaning of the images and the affective information. However, the preference of the diary varied depending on the method applied, suggesting that the scene selection and the description of the diary should be considered simultaneously. \par
Furthermore, we found that the robot's emotion generally improves the preference of the diary regardless of the situation in which the robot describes the emotion. However, the method of presenting emotions still needs to be examined, since describing negative emotions without reasons or presenting mixed emotions at a time decreases the preference of the diary. In addition, the results suggest that there is a relationship between emotions and human-likeness and robot-likeness, that emotions are elements that people may like or dislike differently, and that the contents of the diary should be varied depending on the situation. This research has many implications for the selection method of scenes, contents, and presentation situations, aiming at the establishment of a long-term relationship between humans and robots.\par

\addtolength{\textheight}{-12cm}   % This command serves to balance the column lengths

\bibliographystyle{junsrt}
\bibliography{main}

\begin{thebibliography}{10}

\bibitem{Robohon}
SHARP CORPORATION.
\newblock Robohon.
\newblock \url{https://robohon.com}.

\bibitem{Lovot}
Inc. GROOVE~X.
\newblock Lovot.
\newblock \url{https://help.lovot.life/app/diary/}.

\bibitem{kochigami2021}
Kanae Kochigami, Kei Okada, and Masayuki Inaba.
\newblock Pilot study on robot's open diary to deepen friendships with a child
  and promot communication between a child and people.
\newblock In {\em Companion of the 2021 ACM/IEEE International Conference on
  Human-Robot Interaction}, pp. 104--108, 2021.

\bibitem{Bickmore}
Timothy~W. Bickmore and Rosalind~W. Picard.
\newblock Establishing and maintaining long-term human-computer relationships.
\newblock Vol.~12, No.~2, p. 293^^e2^^80^^93327, jun 2005.

\bibitem{Rivoire2016TheDB}
Claire Rivoire.
\newblock The delicate balance of boring and annoying : Learning proactive
  timing in long-term human robot interaction.
\newblock 2016.

\bibitem{Bruner1986}
Jerome Bruner.
\newblock Actual minds, possible worlds.
\newblock 1986.

\bibitem{Bruner1990}
Jerome Bruner.
\newblock Acts of meaning.
\newblock 1990.

\bibitem{Bruner2002}
Jerome Bruner.
\newblock Making stories:law, literature, life.
\newblock 2002.

\bibitem{doerksen2001source}
Sharon Doerksen and Arthur~P Shimamura.
\newblock Source memory enhancement for emotional words.
\newblock {\em Emotion}, Vol.~1, No.~1, p.~5, 2001.

\bibitem{berntsen2002emotionally}
Dorthe Berntsen and David~C Rubin.
\newblock Emotionally charged autobiographical memories across the life span:
  The recall of happy, sad, traumatic and involuntary memories.
\newblock {\em Psychology and aging}, Vol.~17, No.~4, p. 636, 2002.

\bibitem{buchanan2007retrieval}
Tony~W Buchanan.
\newblock Retrieval of emotional memories.
\newblock {\em Psychological bulletin}, Vol. 133, No.~5, p. 761, 2007.

\bibitem{banaji1994affect}
Mahzarin~R Banaji and Curtis Hardin.
\newblock Affect and memory in retrospective reports.
\newblock {\em Autobiographical memory and the validity of retrospective
  reports}, pp. 71--86, 1994.

\bibitem{waldfogel1948frequency}
Samuel Waldfogel.
\newblock The frequency and affective character of childhood memories.
\newblock {\em Psychological Monographs: General and Applied}, Vol.~62, No.~4,
  p.~i, 1948.

\bibitem{Hunt}
R.~Reed~Hunt.
\newblock {The Concept of Distinctiveness in Memory Research}.
\newblock In {\em {Distinctiveness and Memory}}. Oxford University Press, 04
  2006.

\bibitem{schmidt1991can}
Stephen~R Schmidt.
\newblock Can we have a distinctive theory of memory?
\newblock {\em Memory \& cognition}, Vol.~19, pp. 523--542, 1991.

\bibitem{lewis2017distinctiveness}
Amy Lewis, Josep Call, and Dorthe Berntsen.
\newblock Distinctiveness enhances long-term event memory in non-human
  primates, irrespective of reinforcement.
\newblock {\em American Journal of Primatology}, Vol.~79, No.~8, p. e22665,
  2017.

\bibitem{ichikura2023-1}
Aiko Ichikura, Kento Kawaharazuka, Yoshiki Obinata, Koki Shinjo, Shingo
  Kitagawa, Kei Okada, and Masayuki Inaba.
\newblock Robot's strolling diary -comparison of readers' impressions of
  differences in descriptions in the diary-.
\newblock In {\em ROBOMECH2023}, 2023.
\newblock (to appear).

\bibitem{ichikura2023-2}
Aiko Ichikura, Kento Kawaharazuka, Yoshiki Obinata, Koki Shinjo, Shingo
  Kitagawa, Kei Okada, and Masayuki Inaba.
\newblock Automatic diary generation system including information on joint
  experiences between humans and robots.
\newblock In {\em IAS18}, 2023.
\newblock (to appear).

\bibitem{li2023blip2}
Junnan Li, Dongxu Li, Silvio Savarese, and Steven Hoi.
\newblock Blip-2: Bootstrapping language-image pre-training with frozen image
  encoders and large language models, 2023.

\bibitem{reimers-2019-sentence-bert}
Nils Reimers and Iryna Gurevych.
\newblock Sentence-bert: Sentence embeddings using siamese bert-networks.
\newblock In {\em Proceedings of the 2019 Conference on Empirical Methods in
  Natural Language Processing}. Association for Computational Linguistics, 11
  2019.

\bibitem{brown2020language}
Tom B.~Brown et~al.
\newblock Language models are few-shot learners, 2020.

\end{thebibliography}

\end{document}